\documentclass[10pt,twocolumn,letterpaper]{article}

\usepackage{cvpr}
\usepackage{times}
\usepackage{epsfig}
\usepackage{graphicx}
\usepackage{amsmath}
\usepackage{amssymb}

\usepackage{algorithm}
\usepackage[noend]{algpseudocode}
\usepackage{textcomp} \usepackage{mathtools}
\usepackage[pagebackref=true,breaklinks=true,letterpaper=true,colorlinks,bookmarks=false]{hyperref}

\cvprfinalcopy % *** Uncomment this line for the final submission

\def\cvprPaperID{5734} % *** Enter the CVPR Paper ID here

\ifcvprfinal\fi
\begin{document}

%%%%%%%%% TITLE
\title{Two-Stream Adaptive Graph Convolutional Networks for Skeleton-Based Action Recognition}

\author{
	Lei Shi$^{1,2}$ \and Yifan Zhang$^{1,2}$\thanks{Corresponding Author} \and Jian Cheng$^{1,2,3}$ \and Hanqing Lu$^{1,2}$ \and
	$^1$National Laboratory of Pattern Recognition, Institute of Automation, Chinese Academy of Sciences\\
	$^2$University of Chinese Academy of Sciences\\
	$^3$CAS Center for Excellence in Brain Science and Intelligence Technology\\
	{\tt\small \{lei.shi, yfzhang, jcheng, luhq\}@nlpr.ia.ac.cn} 
}

\maketitle
\thispagestyle{empty}
\begin{abstract}
%         Traditional deep methods for skeleton-based action recognition generally structure the skeleton as a sequence of coordinate vectors or as a pseudo-image to be fed into RNNs or CNNs, which cannot explicitly exploit the natural connectivity among the joints. 
%        Recently, graph convolutional networks (GCNs), which generalize CNNs to more generic non-Euclidean structures, have achieved remarkable performance for skeleton-based action recognition. 
       In skeleton-based action recognition, graph convolutional networks (GCNs), which model the human body skeletons as spatiotemporal graphs, have achieved remarkable performance. 
%        However, the existing GCN-based  methods for constructing graphs generally rely on the fixed and handcraft strategies based on the physical structure of the human body, thus is not optimal for the action recognition task and is difficult to generalize to various data samples. 
       However, in existing GCN-based  methods,  the topology of the graph is set manually, and it is fixed over all layers and input samples. This may not be optimal for the hierarchical GCN  and diverse samples in action recognition  tasks.
       %the strategies for constructing graphs  are generally designed based on the physical structure of the human body, which is not optimal for the action recognition task and is difficult to generalize to various data samples.
       In addition, the second-order information (the lengths and directions of bones) of the skeleton data, which is naturally more informative and discriminative for action recognition, is rarely investigated in existing methods. 
       In this work, we propose a novel two-stream adaptive graph convolutional network (2s-AGCN) for skeleton-based action recognition. 
       The topology of the graph in our model can be either uniformly or individually learned by the BP algorithm in an end-to-end manner. 
       This data-driven method increases the flexibility of the model for graph construction and brings more generality to adapt to various data samples. 
%        Moreover, we exploit the second-order information (the lengths and directions of bones) of the skeleton data for action recognition, which is rarely investigated in existing works. 
       Moreover, a two-stream framework is proposed to model both the first-order and the second-order information simultaneously,  which shows notable improvement for the recognition accuracy.
%        Moreover, previous works only use the first-order information of the skeleton data (the coordinates of joints), whereas the second-order information (the lengths and directions of bones) is less exploited.
%        In this work, a novel two-stream adaptive graph convolutional network (2s-AGCN) is proposed to solve these problems. 
%        The topology of the graph for different layers and samples in the model can be either uniformly or individually learned by the BP algorithm, which provides more flexibility and generality. 
%        Meanwhile, a two-stream framework is proposed to model both the first-order and second-order information simultaneously, which further enhances the recognition performance. 
       Extensive experiments on the two large-scale datasets, NTU-RGBD and Kinetics-Skeleton, demonstrate that the performance of our model exceeds the state-of-the-art with a significant margin.  
    %   The code will be released for future work and to facilitate communication.
	\end{abstract}
    %        This data-driven method provides more flexibility and generality to the model with only a small amount of parameters and additional running-time cost. 

	\section{Introduction}	
    \label{sec:introduction}
    Action recognition methods based on skeleton data have been widely investigated and attracted considerable attention due to their strong adaptability to the dynamic circumstance and complicated background~\cite{vemulapalli_human_2014,fernando_modeling_2015,du_hierarchical_2015,shahroudy_ntu_2016,liu_spatio-temporal_2016,song_end--end_2017,zhang_view_2017,li_skeleton-based_2018,li_independently_2018,liu_two-stream_2017,kim_interpretable_2017,ke_new_2017,liu_enhanced_2017,li_skeleton-based_2017,li_skeleton_2017,yan_spatial_2018,tang_deep_2018,zhang_egogesture:_2018}.
    Conventional deep-learning-based methods manually structure the skeleton as a sequence of joint-coordinate vectors~\cite{du_hierarchical_2015,shahroudy_ntu_2016,liu_spatio-temporal_2016,song_end--end_2017,zhang_view_2017,li_skeleton-based_2018,li_independently_2018} or as a pseudo-image~\cite{liu_two-stream_2017,kim_interpretable_2017,ke_new_2017,liu_enhanced_2017,li_skeleton-based_2017,li_skeleton_2017}, which is  fed into RNNs or CNNs to generate the prediction.
    However, representing the skeleton data as a vector sequence or a 2D grid cannot fully express the dependency between correlated joints. 
    The skeleton is naturally structured as a graph in a non-Euclidean space with the joints as vertexes and their natural connections in the human body as edges. The previous methods cannot exploit the graph structure of the skeleton data and are difficult to generalize to skeletons with arbitrary forms.
    Recently, graph convolutional networks (GCNs), which generalize convolution from image to graph, have been successfully adopted in many applications\cite{kipf_semi-supervised_2016,duvenaud_convolutional_2015,niepert_learning_2016,atwood_diffusion-convolutional_2016,hamilton_inductive_2017,monti_geometric_2017,kipf_neural_2018}. 
    For the skeleton-based action recognition task, Yan et al.~\cite{yan_spatial_2018} first apply GCNs to model the skeleton data. 
    They construct a spatial graph based on the natural connections of joints in the human body and add the temporal edges between corresponding joints in consecutive frames. 
    A distance-based sampling function is proposed for constructing the graph convolutional layer, which is then employed as a basic module to build the final spatiotemporal graph convolutional network (ST-GCN).

%     However, the skeleton graph employed in ST-GCN, which is heuristically predefined, only represents the physical structure of the human body and is not guaranteed to be optimal for various action recognition tasks. 
%     For example, for actions such as ``wiping face" and ``touching head", the relationship between the hands and head is important. 
%     However, it is difficult for ST-GCN to capture the dependency between the hands and head because there are no natural connections in the human body between them and they are located far away from each other in the predefined graph.
%     In addition, the structure of CNNs is hierarchical where different layers contain multiscale semantic information. 
%     Nevertheless, the graph applied in ST-GCN is fixed among all the layers, which lacks the flexibility and capacity to model the multilevel semantic information contained in all of the layers. 
%     Moreover, one graph structure may not fit all action classes.
%     For classes such as ``wiping face" and ``touching head", the connection between the hands and head should be strong, but it is not true for some other classes, such as ``jump up" and ``clapping". However, ST-GCN uses one fixed graph over all action classes.
    
% * <lei.shi@nlpr.ia.ac.cn> 2018-11-15T02:01:42.490Z:
% 
% 这里改成序号类型的怎么样
% 
% ^.
    However, there are three disadvantages for the process of the graph construction in ST-GCN~\cite{yan_spatial_2018}: 
    (1) The skeleton graph employed in ST-GCN is heuristically predefined and represents only the physical structure of the human body. Thus it is not guaranteed to be optimal for the action recognition task. 
    For example, the relationship between the two hands is important for recognizing classes such as ``clapping" and ``reading.''
    However, it is difficult for ST-GCN to capture the dependency between the two hands since they are located far away from each other in the predefined human-body-based graphs. 
    (2) The structure of GCNs is hierarchical where different layers contain multilevel semantic information. However, the topology of the graph applied in ST-GCN is fixed over all the layers, which lacks the flexibility and capacity to model the multilevel semantic information contained in all of the layers; 
    (3) One fixed graph structure may not be optimal for all the samples of different action classes.
    For classes such as ``wiping face" and ``touching head", the connection between the hands and head should be stronger, but it is not true for some other classes, such as ``jumping up" and ``sitting down". 
    This fact suggests that the graph structure should be data dependent, which, however, is not supported in ST-GCN. 
%     However, ST-GCN uses one fixed graph only over all action classes.

    %How to model the diversity of different samples is also a problem in ST-GCN which use only a single graph for all of the samples.
%     For example, it cannot capture the dependency between hands and head as there are no connections naturally in human body. But for many actions such as wiping face and touching head, the relationship between hands and head should be important.

%    	Inspired by Non-local neural networks~\cite{wang_non-local_2017}, 
%     In this work, we propose an adaptive graph convolutional block to solve these problems, in which two types of graphs are trained and updated jointly with other parameters. 
      To solve the above problems, a novel adaptive graph convolutional network is proposed in this work. 
%       , in which two types of graphs are trained and updated jointly with other parameters. 
    It parameterizes two types of graphs, the structure of which are trained and updated jointly with convolutional parameters of the model. 
    One type is a global graph, which represents the common pattern for all the data.
    Another type is an individual graph, which represents the unique pattern for each  data.
%     Moreover, different layers are applied with different graph parameters, 
    Both of the two types of graphs are optimized individually for different layers, which can better fit the hierarchical structure of the model. 
    This data-driven method increases the flexibility of the model for graph construction and brings more generality to adapt to various data samples. 
%     In addition, the adaptive graph module is light weight and very efficient for implementation, adding small amount of parameters and nearly no additional running-time cost. 
%     Besides, different layers are applied with different graph parameters to better fit the hierarchical structure of CNNs. 
%     Moreover, an individual graph will be calculated according to each sample. The overall architecture of the non-local GCN block is shown in Figure~\ref{fig:nlgcn}.
    
    Another notable problem in ST-GCN is that the feature vector attached to each vertex only contains 2D or 3D coordinates of the joints, which can be regarded as the first-order information of the skeleton data. 
    However, the second-order information, which represents the feature of bones between two joints, is not exploited. Typically, the lengths and directions of bones are naturally more informative and discriminative for action recognition.
% * <lei.shi@nlpr.ia.ac.cn> 2018-10-17T01:45:14.601Z:
% 
% 感觉是不是有点生硬强调二阶信息
% 
% ^ <lei.shi@nlpr.ia.ac.cn> 2018-11-14T14:00:28.073Z.
   % Typically, the skeleton data are visualized with both hinged joints and rigid bones. 
  %  In contrast to the joints, the lengths and directions of bones are naturally more obvious and discriminative for observation.
 %   It is nonintuitive to only employ the joint information to classify human activities and ignore the bone information. 
%     To employ the second-order information of the skeletons, the feature of a bone is represented as a vector which contains its length and direction along three dimensions. 
    In order to exploit the second-order information of the skeleton data, the lengths and directions of bones are formulated as a vector pointing from its source joint to its target joint. 
%     In this way, both the length and direction information can be encoded into a single vector and
    Similar to the first-order information, the vector is fed into an adaptive graph convolutional network to predict the action label. 
    Moreover, a two-stream framework is proposed to fuse the first-order and second-order information to further improve the performance. 
    
%     The final model is called the two-stream adaptive graph convolutional network (2s-AGCN). 
    To verify the superiority of the proposed model, namely, the two-stream adaptive graph convolutional network (2s-AGCN),  extensive experiments are performed on two large-scale datasets: NTU-RGBD~\cite{shahroudy_ntu_2016} and Kinetics-Skeleton~\cite{kay_kinetics_2017}. Our model achieves state-of-the-art performance on both of the datasets.
    
%     The main contributions of our work lie in three folds: 
%     \begin{itemize}
%         \item An adaptive graph convolutional network is proposed to adaptively learn the topology of the graph for different GCN layers and skeleton samples in an end-to-end manner, which can better suit the action recognition task and the hierarchical structure of the GCNs.
% %         \item The second-order information, i.e. the lengths and directions of bones, is explicitly formulated and combined with the first-order information, i.e. the joint coordinates, in a two-stream framework to further improve the recognition performance.
%         \item The second-order information of the skeleton data is explicitly formulated and combined with the first-order information using a two-stream framework, which brings notable improvement for the recognition performance.
%         \item On two large-scale datasets for skeleton-based action recognition, the proposed 2s-AGCN exceeds the state-of-the-art by a significant margin. The code will be released for future work and to facilitate communication\footnote{https://github.com/lshiwjx/2s-AGCN}. %our model achieves the state-of-the-art performance.
% 	\end{itemize}    
	    The main contributions of our work lie in three folds: 
	    (1) An adaptive graph convolutional network is proposed to adaptively learn the topology of the graph for different GCN layers and skeleton samples in an end-to-end manner, which can better suit the action recognition task and the hierarchical structure of the GCNs.
        (2) The second-order information of the skeleton data is explicitly formulated and combined with the first-order information using a two-stream framework, which brings notable improvement for the recognition performance.
        (3) On two large-scale datasets for skeleton-based action recognition, the proposed 2s-AGCN exceeds the state-of-the-art by a significant margin. The code will be released for future work and to facilitate communication\footnote{https://github.com/lshiwjx/2s-AGCN}. 
 
	\section{Related work}
    \label{sec:relatedwork}
	\subsection{Skeleton-based action recognition}
    Conventional methods for skeleton-based action recognition usually design handcrafted features to model the human body~\cite{vemulapalli_human_2014,fernando_modeling_2015}.  However, the performance of these handcrafted-feature-based methods is barely satisfactory since it cannot  consider all factors at the same time. 
%     For example, Vemulapalli et al. \cite{vemulapalli_human_2014} encode the skeletons with their rotations and translations in a Lie group. 
%     Fernando et al. \cite{fernando_modeling_2015} leverage the rank pooling method to represent the data with the parameters of the ranker.
    With the development of deep learning, data-driven methods have become the mainstream methods, where the most widely used models are RNNs and CNNs. 
    RNN-based methods usually model the skeleton data as a sequence of the coordinate vectors each represents a human body  joint~\cite{du_hierarchical_2015,shahroudy_ntu_2016,liu_spatio-temporal_2016,song_end--end_2017,zhang_view_2017,li_skeleton-based_2018,li_independently_2018, cao_skeleton-based_2018}. 
%    Du et al. \cite{du_hierarchical_2015} apply a hierarchical bidirectional RNN model to identify the skeleton sequence, which divides the skeleton into different parts and sends them to different subnetworks. 
%     \cite{song_end--end_2017} embeds a spatiotemporal attention module in LSTM based model, so that the network can automatically pay attention to the discriminant spatiotemporal region of the skeleton sequence. 
%    Zhang et al. \cite{zhang_view_2017} introduce the mechanism of view transformation in an LSTM-based model, which automatically translates the skeleton data into a more advantageous angle for action recognition.
    CNN-based methods model the skeleton data as a pseudo-image based on the manually designed transformation rules~\cite{liu_two-stream_2017,kim_interpretable_2017,ke_new_2017,liu_enhanced_2017,li_skeleton-based_2017,li_skeleton_2017}.  The CNN-based methods are generally more popular than RNN-based methods because the CNNs have better parallelizability and are easier to train than RNNs. 
%     Kim et al. \cite{kim_interpretable_2017} apply a one-dimensional residual CNN to identify skeleton sequences where the coordinates of joints are directly concatenated. 
%     \cite{liu_enhanced_2017} manually designs 10 kinds of spatiotemporal images for skeleton encoding, and enhance these images using visual and motion enhancement methods.
%     \cite{liu_two-stream_2017} employs both coordinates and motion information of joints as input, and carefully design a transformer to rearrange the order of joints.
%     Li et al. \cite{li_skeleton_2017} employ multiscale residual networks and various data-augmentation strategies for skeleton-based action recognition.
% * <lei.shi@nlpr.ia.ac.cn> 2018-10-17T01:51:58.258Z:
% 
% 我总觉得简单介绍这些方法是不是没什么意义呢
% 
% ^ <lei.shi@nlpr.ia.ac.cn> 2018-11-14T14:00:34.621Z.
%  the have shown superiority due to their good parallelizability and easier training process compared with RNNs
 
    However, both RNNs and CNNs fail to fully represent the structure of the skeleton data because the skeleton data are naturally embedded in the form of graphs rather than a vector sequence or 2D grids. Recently, Yan et al.~\cite{yan_spatial_2018} propose a spatiotemporal graph convolutional network (ST-GCN) to directly model  the skeleton data as the graph structure. It eliminates the need for designing  handcrafted part assignment or traversal rules, thus achieves better performance than previous methods. 
% Tang et al. \cite{tang_deep_2018} further add the selection of key frames with the help of reinforcement learning.
    
    \subsection{Graph convolutional neural networks}
    There have been many works on graph convolution, and the principle of constructing GCNs mainly follows two streams: spatial perspective and spectral perspective \cite{shuman_emerging_2013,bruna_spectral_2014,henaff_deep_2015,niepert_learning_2016,atwood_diffusion-convolutional_2016,kipf_semi-supervised_2016,duvenaud_convolutional_2015,defferrard_convolutional_2016,hamilton_inductive_2017,monti_geometric_2017,kipf_neural_2018}.
%     Spatial perspective methods consider the convolution as the weighted sum over a regular grid of input feature map. Since the neighbors of each vertex in graph is not same, it need to manually define neighbors adjacent to each vertex \citep{zotero-1824}. 
    Spatial perspective methods directly perform the convolution filters on the graph vertexes and their neighbors, which are extracted and normalized based on manually designed rules \cite{duvenaud_convolutional_2015,niepert_learning_2016,hamilton_inductive_2017,monti_geometric_2017,kipf_neural_2018}.
    In contrast to the spatial perspective methods, spectral perspective methods utilize the eigenvalues and eigenvectors of the graph Laplace matrices. 
    These methods perform the graph convolution in the frequency domain with the help of the graph Fourier transform \cite{shuman_emerging_2013}, which does not need to extract locally connected regions from graphs at each convolutional step \cite{bruna_spectral_2014,henaff_deep_2015,kipf_semi-supervised_2016,defferrard_convolutional_2016}. 
    This work follows the spatial perspective methods.
    
% 	\subsection{Non-local neural networks}
%     The concept of non-local was first proposed in non-local means, which computes a weighted sum of all pixels in an image. It utilize all of the pixels according to their similarity with the center point. The idea is then successfully employed in many other applications. Recently, Wang et al.~\cite{wang_non-local_2017} propose the non-local neural network and achieve the remarkable performance in action recognition area. It present the non-local operations to capture long-range dependencies with deep neural networks, where each response of output feature map is calculated according to all of the features in input feature map.
    
	%\section{Revisit the ST-GCN}
    \section{Graph Convolutional Networks}
    \label{gcn}
%     In this section, we will introduce the ST-GCN. 
    %Due to space restrictions, we only provide a brief introduction in this section. For more details, readers can refer to the original paper~\cite{yan_spatial_2018}.
%     operation and reformulate it with the help of adjacency matrix and directed graph, which makes it more interpretable and extensible. 
%     In Section~\ref{nlgcn}, we present the designed non-local graph convolution block in detail. 
%     In Section~\ref{tsn}, we will describe the methods of utilizing the bone information to further boost the performance. 
    
% 	\subsection{Revisit the ST-GCN}
    \subsection{Graph construction}
\begin{figure}[!htb]
	\centering
	\includegraphics[width=0.9\linewidth]{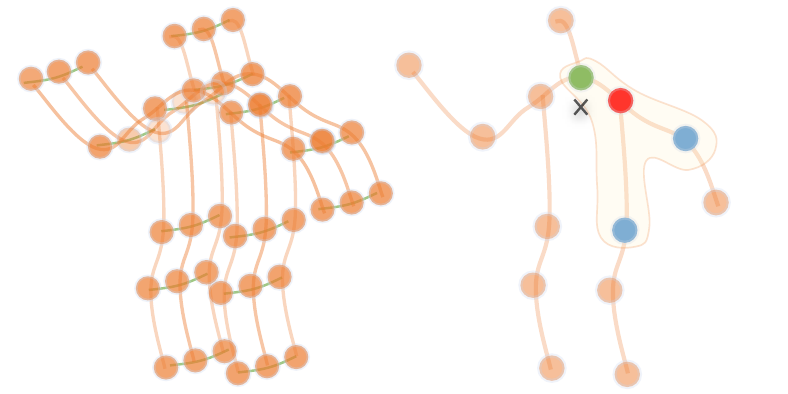}
	\caption{(a).Illustration of the spatiotemporal graph used in ST-GCN. (b).Illustration of the mapping strategy. Different colors denote different subsets.}
	\label{fig:skeleton}
	\end{figure}

%     A skeleton sequence is represented as a list of 2D or 3D coordinate vectors of each human joint in each frame.
    The raw skeleton data in one frame are always provided as a sequence of vectors.
    Each vector represents the 2D or 3D coordinates of the corresponding human joint. 
    A complete action contains multiple frames with different lengths for different samples.
We employ a spatiotemporal graph to model the structured information among these joints along both the spatial and temporal dimensions.
    %Here, spatial dimension refers to the joints in the same frame, and temporal dimension refers to the same joints over all of the frames. 
    The structure of the graph follows the work of ST-GCN~\cite{yan_spatial_2018}.
    The left sketch in Fig.~\ref{fig:skeleton} presents an example of the constructed spatiotemporal skeleton graph, where the joints are represented as vertexes and their natural connections in the human body are represented as spatial edges (the orange lines in Fig.~\ref{fig:skeleton}, left). 
    For the temporal dimension, the corresponding joints between two adjacent frames are connected with temporal edges (the blue lines in Fig.~\ref{fig:skeleton}, left). 
%     The attribute of each vertex is the coordinate vector of the joint. 
    The coordinate vector of each joint is set as the attribute of the corresponding vertex.

    \subsection{Graph convolution}
    Given the graph defined above, multiple layers of spatiotemporal graph convolution operations are applied on the graph to extract the high-level features. 
    The global average pooling layer and the $softmax$ classifier are then employed to predict the action categories based on the extracted features.
%     The constructed graph for each sample is fed into the spatiotemporal graph convolutional neural network(ST-GCN) to predict the action categories. ST-GCN is a stack of 

    In the spatial dimension, the graph convolution operation on vertex $v_i$ is formulated as ~\cite{yan_spatial_2018}:
%     \begin{equation}
%     \label{eq:stgcn}
%     \mathit{f}_{out}(v_{ti})=\sum_{v_{tj}\in \mathcal{B}(v_{ti})} \frac{1}{Z_{ti}(v_{tj}) } \mathit{f}_{in}(v_{tj})\cdot{}\textbf{w}(l_{ti}(v_{tj}))
%     \end{equation}
 	\begin{equation}
    \label{eq:stgcn}
    \mathit{f}_{out}(v_{i})=\sum_{v_{j}\in \mathcal{B}_i} \frac{1}{Z_{ij} } \mathit{f}_{in}(v_{j})\cdot{}w(l_{i}(v_{j}))
    \end{equation}
    where $\mathit{f}$ denotes the feature map and $v$ denotes the vertex of the graph. 
    $\mathcal{B}_i$ denotes the sampling area of the convolution for $v_i$, which is defined as the 1-distance neighbor vertexes ($v_j$) of the target vertex ($v_i$). 
%     As an example, it is shown as colored points in the right sketch of Fig.~\ref{fig:skeleton}. 
%     For example in the right sketch of Fig.~\ref{fig:skeleton}, 1-distance neighbor vertexes of red vertex is painted as colored points.
    $w$ is the weighting function similar to the original convolution operation, which provides a weight vector based on the given input. 
    Note that the number of weight vectors of convolution is fixed, while the number of vertexes in $\mathcal{B}_i$ is varied. 
    To map each vertex with a unique weight vector, a mapping function $l_{i}$ is designed specially in ST-GCN~\cite{yan_spatial_2018}. 
    The right sketch in Fig.~\ref{fig:skeleton} shows this strategy, where $\times$ represents the center of gravity of the skeleton. $\mathcal{B}_i$ is the area enclosed by the curve.
    In detail, the strategy empirically sets the kernel size as 3 and naturally divides $\mathcal{B}_i$ into 3 subsets: 
%     i.e., $\mathcal{S}_{ik}, k=1,2,3$.
    $\mathcal{S}_{i1}$ is the vertex itself (the red circle in Fig.~\ref{fig:skeleton}, right); 
    $\mathcal{S}_{i2}$ is the centripetal subset, which contains the neighboring vertexes that are closer to the center of gravity (the green circle); 
    $\mathcal{S}_{i3}$ is the centrifugal subset, which contains the neighboring vertexes that are farther from the center of gravity (the blue circle). 
    $Z_{ij}$ denotes the cardinality of $\mathcal{S}_{ik}$ that contains $v_j$. It aims to balance the contribution of each subset.
%     \textbf{(The description of $Z_{i}(v_{j})$ is wrong.  Look at paper~\cite{yan_spatial_2018})}
%     The right sketch of Fig.~\ref{fig:skeleton} shows this strategy, where different colors represent different subsets and each color is corresponded with the individual learnable weight vector.

    \subsection{Implementation}
    The implementation of the graph convolution in the spatial dimension is not straightforward.
    Concretely, the feature map of the network is actually a $C\times T\times N$ tensor, where $N$ denotes the number of vertexes, $T$ denotes the temporal length and $C$ denotes the number of channels. 
    To implement the ST-GCN, Eq.~\ref{eq:stgcn} is transformed into
    \begin{equation}
    \label{eq:stgcni}
    \mathbf{f}_{out} = \sum_k^{K_v} \mathbf{W}_k (\mathbf{f}_{in} {\mathbf{A}}_k) \odot\mathbf{M}_k
    \end{equation}
%     Here, $\mathbf{f}$ is the $C\times T\times N$ feature map where $N$ denotes the number of vertexes, $T$ denotes the temporal length and $C$ denotes the number of channels. 
where $K_v$ denotes the kernel size of the spatial dimension. With the partition strategy designed above, $K_v$ is set to $3$. 
    $\mathbf{A}_k = {\mathbf{\Lambda}}_k^{-\frac{1}{2}} {\mathbf{\bar{A}}}_k {\mathbf{\Lambda}}_{k}^{-\frac{1}{2}}$, where $\mathbf{\bar{A}}_k$ is similar to the $N\times N$ adjacency matrix, and its element $\mathbf{\bar{A}}_k^{ij}$ indicates whether the vertex $v_j$ is in the subset $S_{ik}$ of vertex $v_i$. 
    It is used to extract the connected vertexes in a particular subset from $\mathbf{f}_{in}$ for the corresponding weight vector.
    $\mathbf{\Lambda}_k^{ii}=\sum_j (\mathbf{\bar{A}}_k^{ij})+\alpha$ is the normalized diagonal matrix. 
    $\alpha$ is set to $0.001$ to avoid empty rows. 
%     $\mathbf{A_0}=\mathbf{I}$, which denotes the self-connections of vertexes.
%     $\mathbf{A_1}$ denotes the connections of centripetal subset and $\mathbf{A_2}$ denotes the centrifugal subset. 
    $\mathbf{W}_k$ is the $C_{out}\times C_{in}\times 1 \times 1$ weight vector of the $1\times 1$ convolution operation, which represents the weighting function $w$ in Eq.~\ref{eq:stgcn}. 
    $\mathbf{M}_k$ is an $N\times N$ attention map that indicates the importance of each vertex. 
    $\odot$ denotes the dot product.
%     , which means it can only effect the vertexes that are connected with current target.
    
    For the temporal dimension, since the number of neighbors for each vertex is fixed as $2$ (corresponding joints in the two consecutive frames), it is straightforward to perform the graph convolution similar to the classical convolution operation. Concretely, we perform a $K_t\times 1$ convolution on the output feature map calculated above, where $K_t$ is the kernel size of temporal dimension.

% * <yfzhang@nlpr.ia.ac.cn> 2018-11-06T03:36:08.031Z:
% 
% K_t is not defined
% 
% ^ <yfzhang@nlpr.ia.ac.cn> 2018-11-06T03:37:21.250Z.
    \section{Two-stream adaptive graph convolutional network}
    In this section, we introduce the components of our proposed two-stream adaptive graph convolutional network (2s-AGCN) in detail. 
	\subsection{Adaptive graph convolutional layer}
    \label{nlgcn}
%     The spatiotemporal graph convolution for skeleton data described above is calculated based on the given graph $\mathcal{G}$, which is manually designed according to the natural connections of human body. 
%     However, it is hard to say this structure is the best choice for action recognition. 
%     For example, there is no connection between hand and head in the official provided graph of NTU-RGBD dataset. But for many actions such as wiping face and touching head, the relationship between hand and head should be important. 
%     So, the connection relationship should not be constrained in the adjacent joints. 
%     Besides, as the information is transferred from lower layers to higher layers, the semantic information contained in different layers is also varied. One shared graph structure can not adapt to this variety. The structure of graph should be updated along the message passing. Moreover, due to the deformation of skeleton for different classes, relations among joints in different samples should also be different. 

%     To solve the problem, inspired by the non-local neural network, we propose the non-local graph convolution which directly focus on all of the joints to decide whether there are connections between pairs of vertexes. 
%     The graph structure is learned individually for different layers and samples in the training process in an end-to-end manner. 
%     Different with the original non-local block, our non-local graph convolution block contains three parts.
	The spatiotemporal graph convolution for the skeleton data described above is calculated based on a predefined graph, which may not be the best choice as explained in Sec.~\ref{sec:introduction}.
% 	As explained in introduction, the predefined graph, 
%     Firstly, it is constructed according to the natural connections of human body, which is not guaranteed to be suitable for action recognition task. 
%     Besides, it is shared among all the convolutional layers, but the structure of CNN is hierachical where different layers contain different-level semantic information.
%     Thirdly, the graphs should be data dependent since different samples may need different graphs. 
    To solve this problem, we propose an adaptive graph convolutional layer. 
    It makes the topology of the graph optimized together with the other parameters of the network in an end-to-end learning manner. 
    The graph is unique for different layers and samples, which greatly increases the flexibility of the model. 
    Meanwhile, it is designed as a residual branch, which guarantees the stability of the original model.
%     meanwhile guarantees the stability of model with a residual connection.

    In detail, according to Eq.~\ref{eq:stgcni}, the topology of the graph is actually decided by the adjacency matrix and the mask, i.e., $\mathbf{A}_k$ and  $\mathbf{M}_k$, respectively. 
    $\mathbf{A}_k$ determines whether there are connections between two vertexes and $\mathbf{M}_k$ determines the strength of the connections. 
%     To make the graph structure adaptively, we divide $\mathbf{A}_i$ into three parts.
%     Fig.~\ref{fig:nlgcn} shows the structure of the basic non-local graph convolution block. 
%     similar to Eq.~\ref{eq:stgcni}, it can be written as:
    To make the graph structure adaptive, we change Eq.~\ref{eq:stgcni} into the following form:
    \begin{equation}
    \label{eq:nlstgcn}
    \mathbf{f}_{out} = \sum_k^{K_v} \mathbf{W}_k \mathbf{f}_{in} ({\mathbf{A}_{k}}+{\mathbf{B}_{k}}+{\mathbf{C}_{k}})
%     + \mathbf{W}_{res} \mathbf{f}_{in}
    \end{equation}
    %     \begin{equation}
    % \label{eq:nlstgcn}
    % \mathbf{f}_{out} = \sum_k^{K_v} \mathbf{W}_k \mathbf{f}_{in} ({\mathbf{A}_{k}}+{\mathbf{B}_{k}})
%     + \mathbf{W}_{res} \mathbf{f}_{in}
    % \end{equation}
    The main difference lies in the adjacency matrix of the graph, which is divided into three parts: $\mathbf{A}_k$, $\mathbf{B}_k$ and $\mathbf{C}_k$.
%     It contains three parts.
%     The first part is the physical graph ($A_i$ in Fig.~\ref{fig:nlgcn}) same as ST-GCN. It represents the natural connections of the human body.

%      \textbf{The first part} is a $N\times N$ adjacency matrix ($A_i$ in Fig.~\ref{fig:nlgcn}), which represents the natural connections of the human body. It is the same as the $A_i$ in Eq.~\ref{eq:stgcni}
     \textbf{The first part ($\mathbf{A}_k$)} is the same as the original normalized $N\times N$ adjacency matrix $\mathbf{A}_k$ in Eq.~\ref{eq:stgcni}. It represents the physical structure of the human body.
 
%     The second part is a shared graph ($B_i$) which is same for different samples. 
    \textbf{The second part ($\mathbf{B}_k$)} is also an $N\times N$ adjacency matrix. In contrast to $\mathbf{A}_k$, the elements of $\mathbf{B}_k$ are parameterized and optimized together with the other parameters in the training process.
%     The value of each element denotes not only whether there exist connections between two joints, but also the strength of the connections. 
%     It can represent the common pattern of connections between joints. 
	There are no constraints on the value of $\mathbf{B}_k$, which means that the graph is completely learned according to the training data. With this data-driven manner, the model can learn graphs that are fully targeted to the recognition task and more individualized for different information contained in different layers.
%     With this data-driven manner, the model can learn graphs that are more suitable for recognition task based on the data and given labels. 
%     Besides, each layer contains their own graph parameters, which means different layers can learn different graphs since they contain different-level semantic information.
    Note that the element in the matrix can be an arbitrary value. It indicates not only the existence of the connections between two joints but also the strength of the connections. 
    It can play the same role of the attention mechanism performed by $\mathbf{M}_k$ in Eq.~\ref{eq:stgcni}%~\cite{yan_spatial_2018}.
% * <lei.shi@nlpr.ia.ac.cn> 2018-10-17T02:10:03.751Z:
% 
% 这段语法总觉得不流畅
% 
% ^ <lei.shi@nlpr.ia.ac.cn> 2018-11-14T14:00:55.440Z.
%     but is more flexible.
	However, the original attention matrix $\mathbf{M}_k$ is dot multiplied to $A_k$, which means that if one of the elements in $A_k$ is $0$, it will always be $0$ irrespective the value of $\mathbf{M}_k$. Thus, it cannot generate the new connections that not exist in the original physical graph. 
    From this perspective, $\mathbf{B}_k$ is more flexible than $\mathbf{M}_k$.
%     In this way, there is no need to multiply $\mathbf{M}_i$ anymore. 
%     To avoid the redundant parameter, we remove the $\mathbf{M}_i$. 
%     Just as predicted, it does not affect the performance.
%     We also tried adding some constrains on the elements of the matrix, but found no improvement.
    
    \textbf{The third part} ($\mathbf{C}_{k}$) is a data-dependent graph which learn a unique graph for each sample. 
%     It will capture the unique features of each sample. 
%     To decide whether there is a connection between two vertexes and how strong is the connection, we first embed the vertexes into another space with the embedding function $\theta$ and $\phi$. It can also reduce the dimension of the vertexes to save the calculation. Here, with expensive experiments, we choose the single fully connected layer as the embedding function, which reduce the number of channels to the half of the original number of channels.
    To determine whether there is a connection between two vertexes and how strong the connection is, we apply the normalized embedded Gaussian function to calculate the similarity of the two vertexes:
    \begin{equation}
    \label{eq:gaussian}
    \mathit{f}(v_i,v_j)=\frac{e^{\theta(v_i)^T\phi(v_j)}}{\sum_{j=1}^N{e^{\theta(v_i)^T\phi(v_j)}}}
    \end{equation}
    where $N$ is the total number of the vertexes. We use the dot product to measure the similarity of the two vertexes in an embedding space.
    In detail, given the input feature map $\mathbf{f_{in}}$ whose size is $C_{in}\times T \times N$, we first embed it into  $C_{e}\times T \times N$ with two embedding functions, i.e., $\theta$ and $\phi$. 
    Here, through extensive experiments, we choose one $1\times 1$ convolutional layer as the embedding function. 
    The two embedded feature maps are rearranged and reshaped to an $N\times C_{e}T$ matrix and a $C_{e}T \times N$ matrix. They are then multiplied to obtain an $N \times N$ similarity matrix $\mathbf{C}_k$, whose element $\mathbf{C}_k^{ij}$ represents the similarity of vertex $v_i$ and vertex $v_j$. The value of the matrix is normalized to $0-1$, which is used as the soft edge of the two vertexes.
%     Since the normalized Gaussian is equipped with $softmax$ operation, Eq.\ref{eq:gaussian} can be transformed into:
    Since the normalized Gaussian is equipped with a $softmax$ operation, we can calculate $\mathbf{C}_k$ based on Eq.\ref{eq:gaussian} as follows:
%     For individual graph, we apply the embedded Gaussian function to calculate the similarity of two joints:
% 	Then, we apply the Gaussian function to calculate the similarity of two embedded vertexes and use the similarity as the connections between vertexes. We add a soft
%     \begin{equation}
%     \mathit{f}(v_i,v_j)=e^{\theta(v_i)^T\phi(v_j)}
%     \end{equation}
%     using $1\times 1$ convolution to represent the embedding functions, the individual graph for each input feature map is calculated by:
    \begin{equation}
    \mathbf{C}_k = softmax(\mathbf{f_{in}}^T\mathbf{W}^T_{\theta k}\mathbf{W}_{\phi k} \mathbf{f_{in}})
    \end{equation}
        % \begin{equation}
    % \mathbf{B}_k = softmax(\mathbf{f_{in}}^T\mathbf{W}^T_{\theta k}\mathbf{W}_{\phi k} \mathbf{f_{in}})
    % \end{equation}
%     where the $softmax$ operation normalizes the result of product to $0-1$. 
%     where $\mathbf{W}$ is the parameters of the embedding function and is initialed with 0. 
    where $\mathbf{W_\theta}$ and $\mathbf{W_\phi}$ are the parameters of the embedding functions $\theta$ and $\phi$, respectively. 
%     Same as the shared graph, it is also employed as a residual connection. 
    
%     All of the three parts are important, which is verified in the ablation study in Section~\ref{sec:ablation}.
    
    \begin{figure}[tb]
	\begin{center}
	\includegraphics[width=0.95\linewidth]{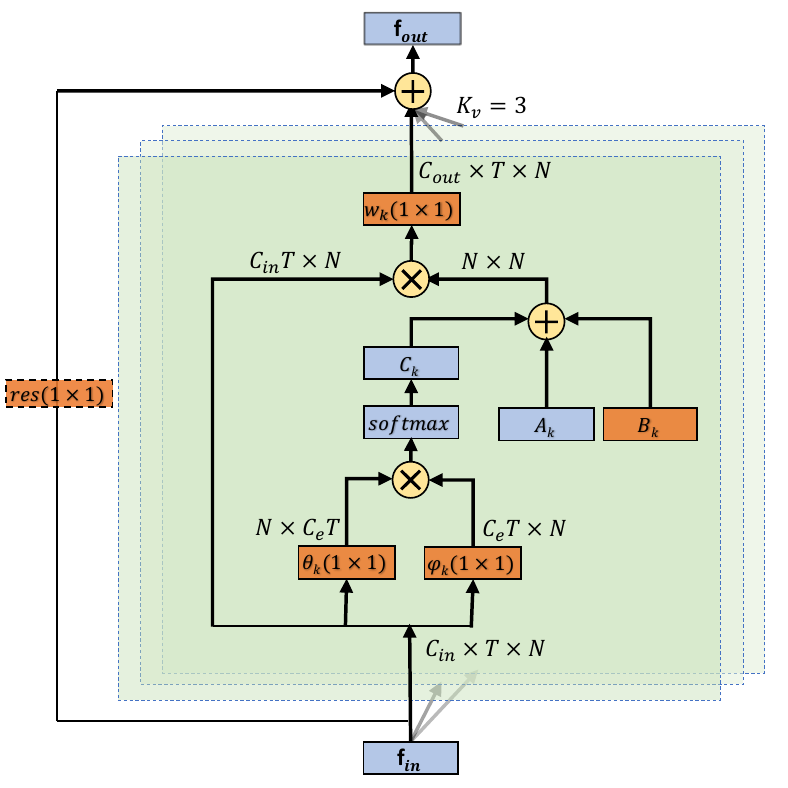}
	\caption{Illustration of the adaptive graph convolutional layer. There are a total of three types of graphs in each layer, i.e., $\mathbf{A}_k$, $\mathbf{B}_k$ and $\mathbf{C}_k$. The orange box indicates that the parameter is learnable. $(1\times 1)$ denotes the kernel size of convolution. $K_v$ denotes the number of subsets. $\oplus$ denotes the elementwise summation. $\otimes$ denotes the matirx multiplication. The residual box (dotted line) is only needed when $C_{in}$ is not the same as $C_{out}$.}
	\label{fig:nlgcn}	
	\end{center}
	\end{figure}
    
% * <lei.shi@nlpr.ia.ac.cn> 2018-10-17T02:12:06.120Z:
% 
% 这个图是不是可以放在文章开头
% 
% ^ <lei.shi@nlpr.ia.ac.cn> 2018-11-14T14:00:59.465Z.
%     In detail, according to the Eq.~\ref{eq:stgcni}, the structure of graph is actually decided by $\mathbf{A}_i$. 
%     For shared graph, we add $3$ $N\times N$ learnable parameter $\mathbf{B}_i$ to represent the connections between each pair of joints. 
%     The value of each element denote not only whether there exist edge between two joints, but also the strength of the connection or the similarly of the two joints. 
    Rather than directly replacing the original $\mathbf{A}_k$ with $\mathbf{B}_k$ or $\mathbf{C}_k$, we add them to it. %$\mathbf{A}_k$. 
%     The value of $\mathbf{B}_k$ is initialed with 0. 
%     The initial value of $\mathbf{C}_i$ is also $0$ since the parameter of embedding function is initialized with $0$. 
    The value of $\mathbf{B}_k$ and the parameters of $\theta$ and $\phi$ are initialized to $0$. 
%     which guarantees the initial value of $\mathbf{C}_k$ is also $0$.
    In this way, it can strengthen the flexibility of the model without degrading the original performance. 
%     All of the three parts are important, which is verified in the ablation study in Section~\ref{sec:ablation}.
%     Note that it can also play the role of attention mechanism performed by $\mathbf{M}_i$ in Eq.~\ref{eq:stgcni} but is more flexible, because the element with $0$ will always be $0$ no matter what $\mathbf{M}_i$ is. We remove the $\mathbf{M}_i$ to avoid the redundant parameter and found it does not affect the performance.
    
%     For individual graph, we apply the embedded Gaussian function to calculate the similarity of two joints:
%     \begin{equation}
%     \mathit{f}(v_i,v_j)=e^{\theta(v_i)^T\phi(v_j)}
%     \end{equation}
%     using $1\times 1$ convolution to represent the embedding functions, the individual graph for each input feature map is calculated by:
%     \begin{equation}
%     \mathbf{C}_j = softmax(\mathbf{f_{in}}^T\mathbf{W}^T_{\theta j}\mathbf{W}_{\phi j} \mathbf{f_{in}})
%     \end{equation}
%     where the $softmax$ operation normalizes the result of product to $0-1$. $\mathbf{W}$ is the parameters of the $1\times 1$ convolution and is initialed with 0. Same as the shared graph, it is also employed as a residual connection. Thus the Eq.~\ref{eq:stgcni} becomes:
    
    The overall architecture of the adaptive graph convolution layer is shown in Fig.~\ref{fig:nlgcn}. 
    Except for the $\mathbf{A}_k$, $\mathbf{B}_k$ and $\mathbf{C}_k$ introduced above, the kernel size of convolution ($K_v$) is set the same as before, i.e., 3. 
    $w_k$ is the weighting function introduced in Eq.~\ref{eq:stgcn}, whose parameter is $\mathbf{W}_k$ in Eq.~\ref{eq:nlstgcn}. 
%     We also test $K_v>3$ and found no improvement.
    A residual connection, similar to \cite{he_deep_2016}, is added for each layer, which allows the layer to be inserted into any existing models without breaking its initial behavior. If the number of input channels is different than the number of output channels, a $1\times 1$ convolution (orange box with dashed line in Fig.~\ref{fig:nlgcn}) is inserted in the residual path to transform the input to match the output in the channel dimension.
%     All of the changes are effective and efficient.
   
    \subsection{Adaptive graph convolutional block}
\begin{figure}[!htb]
	\begin{center}
	\includegraphics[width=0.8\linewidth]{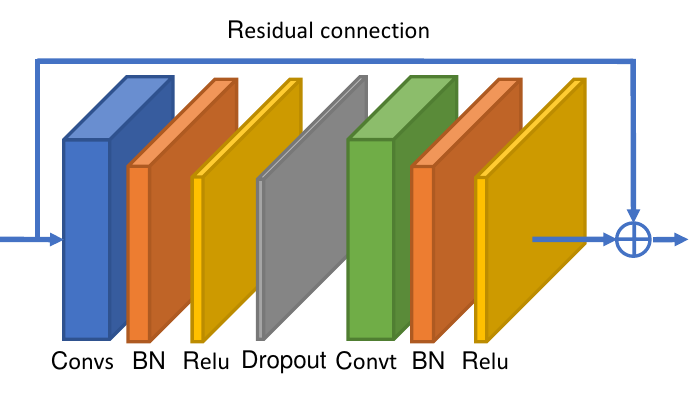}
	\caption{Illustration of the adaptive graph convolutional block. Convs represents the spatial GCN, and Convt represents the temporal GCN, both of which are followed by a BN layer and a ReLU layer. Moreover, a residual connection is added for each block.}
	\label{fig:block}	
	\end{center}
	\end{figure}

    The convolution for the temporal dimension is the same as ST-GCN, i.e., performing the $K_t\times 1$ convolution on the $C\times T \times N$ feature maps. 
    Both the spatial GCN and temporal GCN are followed by a batch normalization (BN) layer and a ReLU layer. 
    As shown in Fig.~\ref{fig:block}, one basic block is the combination of one spatial GCN (Convs), one temporal GCN (Convt) and an additional dropout layer with the drop rate set as 0.5. 
    To stabilize the training, a residual connection is added for each block.      
    
	\subsection{Adaptive graph convolutional network}
\begin{figure}[!htb]
	\begin{center}
	\includegraphics[width=0.8\linewidth]{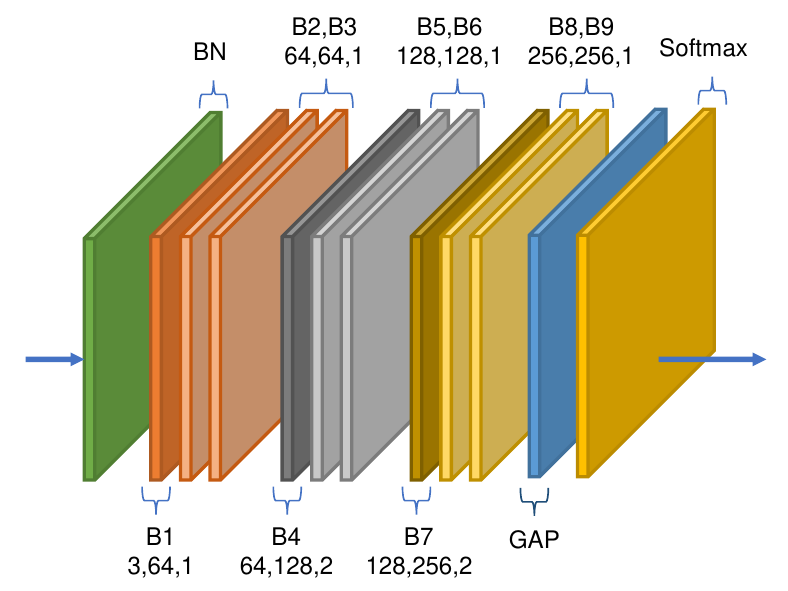}
	\caption{Illustration of the AGCN. There are a total of $9$ blocks (B1-B9). The three numbers of each block represent the number of input channels, the number of output channels and the stride, respectively. GAP represents the global average pooling layer.}
	\label{fig:stream}	
	\end{center}
	\end{figure}

    The adaptive graph convolutional network (AGCN) is the stack of these basic blocks, as shown in Fig.~\ref{fig:stream}. There are a total of $9$ blocks. 
    The numbers of output channels for each block are $64$, $64$, $64$, $128$, $128$, $128$, $256$, $256$ and $256$. 
    A data BN layer is added at the beginning to normalize the input data.
    A global average pooling layer is performed at the end to pool feature maps of different samples to the same size. 
    The final output is sent to a $softmax$ classifier to obtain the prediction. 
%     More details about the model can be found in the code, which will be released afterwards.
    
    \subsection{Two-stream networks}
    \label{tsn}
    \begin{figure*}[!htb]
	\begin{center}
	\includegraphics[width=0.8\linewidth]{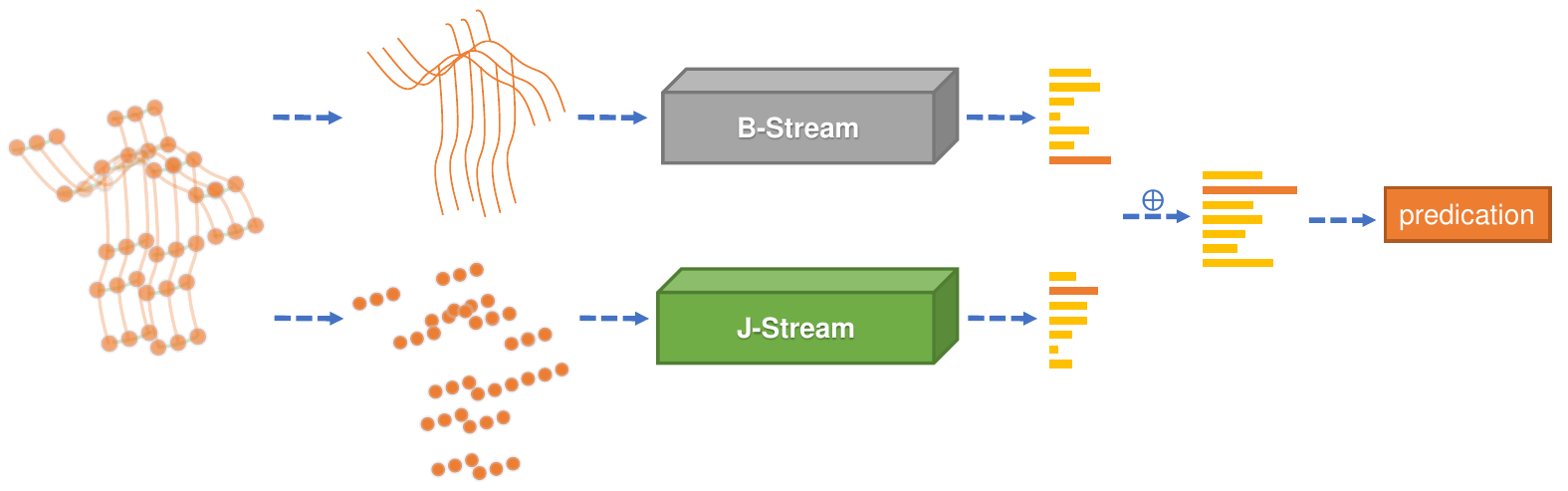}
	\caption{Illustration of the overall architecture of the 2s-AGCN. The scores of two streams are added to obtain the final prediction.}
	\label{fig:twostream}	
	\end{center}
	\end{figure*}

%     Traditional methods only use coordinates of body joints as input, which have three channels along $x$, $y$ and $z$ axises. 
%     It is the first-order information which only relies on the single joint. 
%     However, the second-order information, which represents the bones between joints, is also important for recognition but is neglected or not emphasized. 
%     In this sense, we propose to explicitly model the second-order information, the bone information, with another stream to boost the action recognition. 
    As introduced in Sec.~\ref{sec:introduction}, the second-order information, i.e., the bone information, is also important for skeleton-based action recognition but is neglected in previous works.
    In this paper, we propose explicitly modeling the second-order information, namely, the bone information, with a two-stream framework to enhance the recognition. 
    
%     In particular, we represent each bone as a vector pointing from its source joint to its tail joint, which contains not only the length information, but also the direction information.
%     The source joint is defined as the joint that are closer to the gravity center of the skeleton and the tail joint is the joint that are further to the gravity center. 
    In particular, since each bone is bound with two joints, we define that the joint close to the  center of gravity of the skeleton is the source joint and the joint far away from the center of gravity is the target joint. 
    Each bone is represented as a vector pointing to its target joint from its source joint, which contains not only the length information, but also the direction information.
    For example, given a bone with its source joint $\mathbf{v_1}=(x_1, y_1, z_1)$ and its target joint $\mathbf{v_2}=(x_2, y_2, z_2)$, the vector of the bone is calculated as $\mathbf{e_{v_1,v_2}} = (x_2-x_1, y_2-y_1, z_2-z_1)$.
%     In particular, the input graph of the new stream, which we call B-stream, has the same topological structure with the graph of original stream (J-stream). 
%     Each vertex of B-stream has six channels, which contain the length and direction of the bone at the source of the current joint. 
%     Each vertex of B-stream is a vector which represents the length and direction of the bone at the source of the current joint. 

    Since the graph of the skeleton data have no cycles, each bone can be assigned with a unique target joint.
    The number of joints is one more than the number of bones because the central joint is not assigned to any bones. 
    To simplify the design of the network, we add an empty bone with its value as $0$ to the central joint. 
    In this way, both the graph and the network of bones can be designed the same as that of joints because each bone can be bound with a unique joint. 
    We use J-stream and B-stream to represent the networks of joints and bones, respectively. 
%     except for the central joint, which will be filled with 0. 
%     The bone vector is calculated by the difference between coordinates of two body joints along the $x$, $y$ and $z$ axises.
    The overall architecture (2s-AGCN) is shown in Fig.~\ref{fig:twostream}. 
    Given a sample, we first calculate the data of bones based on the data of joints. Then, the joint data and bone data are fed into the J-stream and B-stream, respectively.
    Finally, the $softmax$ scores of the two streams are added to obtain the fused score and predict the action label. 
%     Other fusion methods are left as the future work.

	\section{Experiments}
\begin{figure}[!htb]
	\begin{center}
	\includegraphics[width=0.8\linewidth]{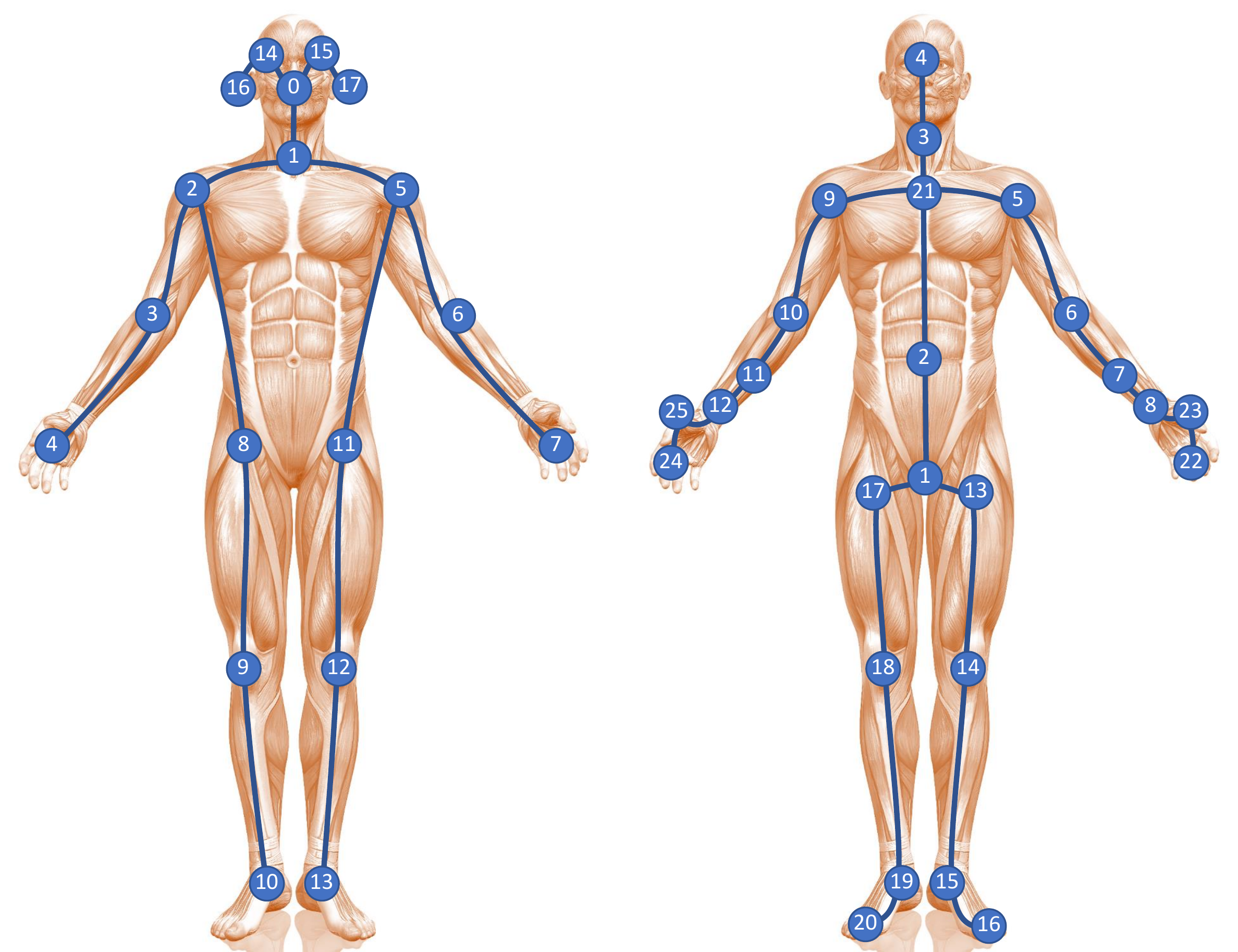}
	\caption{The left sketch shows the joint label of the Kinetics-Skeleton dataset and the right sketch shows the joint label of the NTU-RGBD dataset.}
	\label{fig:dataset}	
	\end{center}
	\end{figure}

    To perform a head-to-head comparison with ST-GCN, our experiments are conducted on the same two large-scale action recognition datasets: NTU-RGBD \cite{shahroudy_ntu_2016} and Kinetics-Skeleton \cite{kay_kinetics_2017,yan_spatial_2018}. 
    First, since the  NTU-RGBD dataset is smaller than the Kinetics-Skeleton dataset, we perform exhaustive ablation studies on it to examine the contributions of the proposed model components based on the recognition performance. 
    Then, the final model is evaluated on both of the datasets to verify the generality and is compared with the other state-of-the-art approaches. 
    The definitions of joints and their natural connections in the two datasets are shown in Fig.~\ref{fig:dataset}.
    
    \subsection{Datasets}
    \textbf{NTU-RGBD:} NTU-RGBD~\cite{shahroudy_ntu_2016} is currently the largest and most widely used in-door-captured action recognition dataset, which contains 56,000 action clips in 60 action classes. The clips are performed by 40 volunteers in different age groups ranging from 10 to 35. Each action is captured by 3 cameras at the same height but from different horizontal angles: $-45^\circ, 0^\circ, 45^\circ$. 
    This dataset provides 3D joint locations of each frame detected by Kinect depth sensors. There are 25 joints for each subject in the skeleton sequences, while each video has no more than 2 subjects. The original paper \cite{shahroudy_ntu_2016} of the dataset recommends two benchmarks: 
    1). Cross-subject (X-Sub): the dataset in this benchmark is divided into a training set (40,320 videos) and a validation set (16,560 videos), where the actors in the two subsets are different.
    2).Cross-view (X-View): the training set in this benchmark contains 37,920 videos that are captured by cameras 2 and 3, and the validation set contains 18,960 videos that are captured by camera 1.
    We follow this convention and report the top-1 accuracy on both benchmarks.
    
    \textbf{Kinetics-Skeleton:} Kinetics \cite{kay_kinetics_2017} is a large-scale human action dataset that contains 300,000 videos clips in 400 classes. The video clips are sourced from YouTube videos and have a great variety. It only provides raw video clips without skeleton data. 
    \cite{yan_spatial_2018} estimate the locations of 18 joints on every frame of the clips using the publicly available OpenPose toolbox \cite{cao_realtime_2017}. Two peoples are selected for multiperson clips based on the average joint confidence. 
    We use their released data (Kinetics-Skeleton) to evaluate our model. 
    The dataset is divided into a training set (240,000 clips) and a validation set (20,000 clips). Following the evaluation method in \cite{yan_spatial_2018}, we train the models on the training set and report the top-1 and top-5 accuracies on the validation set.
	
    \subsection{Training details}
     All experiments are conducted on the PyTorch deep learning framework \cite{paszke_automatic_2017}.
%      with 8 TITANX GPUs. 
     Stochastic gradient descent (SGD) with Nesterov momentum ($0.9$) is applied as the optimization strategy. The batch size is $64$. Cross-entropy is selected as the loss function to backpropagate gradients. The weight decay is set to $0.0001$.
     
    For the NTU-RGBD dataset, there are at most two people in each sample of the dataset. If the number of bodies in the sample is less than 2, we pad the second body with 0. The max number of frames in each sample is 300. 
    For samples with less than 300 frames, we repeat the samples until it reaches 300 frames. The learning rate is set as $0.1$ and is divided by $10$ at the $30_{th}$ epoch and $40_{th}$ epoch. The training process is ended at the $50_{th}$ epoch. 
     
    For the Kinetics-Skeleton dataset, the size of the input tensor of Kinetics is set the same as \cite{yan_spatial_2018}, which contains 150 frames with 2 bodies in each frame. 
    We perform the same data-augmentation methods as in \cite{yan_spatial_2018}.
%     We do not apply any data preprocessing in Kinetics but perform some data-augmentation methods similar to \cite{yan_spatial_2018}. 
    In detail, we randomly choose 150 frames from the input skeleton sequence and slightly disturb the joint coordinates with randomly chosen rotations and translations. The learning rate is also set as $0.1$ and is divided by $10$ at the $45_{th}$ epoch and $55_{th}$ epoch. The training process is ended at the $65_{th}$ epoch. 
     
    \subsection{Ablation Study}
    \label{sec:ablation}
    We examine the effectiveness of the proposed components in two-stream adaptive graph convolutional network (2s-AGCN) in this section with the X-View benchmark on the NTU-RGBD dataset. The original performance of ST-GCN on the NTU-RGBD dataset is $88.3\%$. By using the rearranged learning-rate scheduler and the specially designed data preprocessing methods, it is improved to $92.7\%$, which is used as the baseline in the experiments. 
    The detail is introduced in the supplementary material.

    \subsubsection{Adaptive graph convolutional block.}
    As introduced in Section~\ref{nlgcn}, there are $3$ types of graphs in the adaptive graph convolutional block, i.e., $A$, $B$ and $C$. 
    We manually delete one of the graphs and show their performance in Tab.~\ref{tab:nlgcn}. 
    This table shows that adaptively learning the graph is beneficial for action recognition and that deleting any one of the three graphs will harm the performance. 
    With all three graphs added together, the model obtains the best performance. 
    We also test the importance of $\mathbf{M}$ used in the original ST-GCN. 
    The result shows that given each connection, a weight parameter is important, which also proves the importance of the adaptive graph structure.
%     Table~\ref{adaptive} shows that adaptively learning the graph is benefit for action recognition. This is intuitive because the manually designed graph is hard to meet the requirements of action recognition. Besides, it shows that more flexibility will bring better performance, where the performance of " Learning $P_k$ for each layer " is higher than "Learning $P_k$" and "Learning $P_k$" is better than "Learning $L$ ". The best performance is gained when applying unique parameters for each convolutional layer. It confirms our conjecture that different convolutional layers contain different semantic information, where a shared graph is not enough to model the diversities.
    
    %L strategy means only parameterize the Laplacian matrix. A means parameterize all the $L^k, k=0,1,\cdots,K$. In other words, the neighbors of vertex in each convolution step is adaptively learned by model in the training process. Share and individual indicate whether the parameter is shared by all convolution layer or not. 
	\begin{table}[htb]
 	\begin{center}
		\begin{tabular}{lc}
			\hline
			Methods     & Accuracy ($\%$)\\
			\hline
            ST-GCN      & 92.7      \\   
            ST-GCN wo/M & 91.1     \\
            AGCN wo/A  & 93.4         \\
            AGCN wo/B  & 93.3       \\
			AGCN wo/C  & 93.4      \\ 
            AGCN       & 93.7         \\  
			\hline
		\end{tabular}
        \end{center}
     	\caption{Comparisons of the validation accuracy when adding adaptive graph convolutional block with or without $A$, $B$ and $C$. wo/X means deleting the X module. }
        \label{tab:nlgcn}
	\end{table}
    
    \subsubsection{Visualization of the learned graphs}
\begin{figure}[!htb]
	\begin{center}
	\includegraphics[width=1\linewidth]{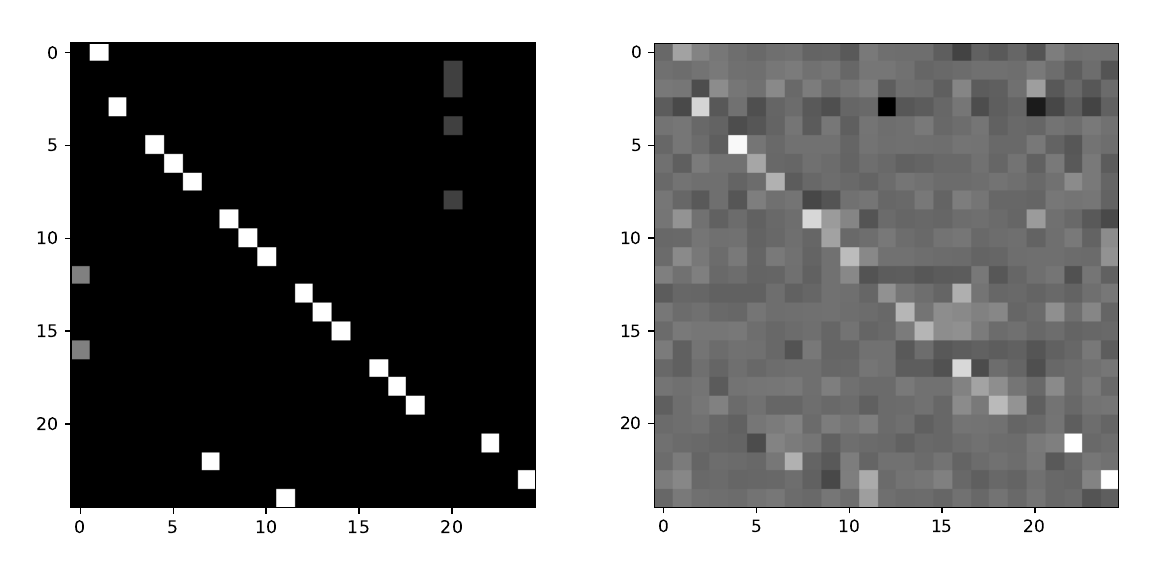}
	\caption{Example of the learned adjacency matrix. The left matrix is the original adjacency matrix for the second subset in the NTU-RGBD dataset. The right matrix is an example of the corresponding adaptive adjacency matrix learned by our model. 
%     The color of each element $A_{ij}$ in matrix represents the tightness of the connection between joint $i$ and joint $j$.
    }
	\label{fig:map}	
	\end{center}
	\end{figure}

    Fig.~\ref{fig:map} shows an example of the adjacency matrix learned by our model for the second subset.
	The gray scale of each element in the matrix represents the strength of the connection.
%     The left is the original adjacency matrix for second subset employed in ST-GCN, which is same for different layers and different samples. 
    The left is the original adjacency matrix for the second subset employed in ST-GCN, and the right is an example of the corresponding adaptive adjacency matrix learned by our model. 
    It is clear that the learned structure of the graph is more flexible and not constrained to the physical connections of the human body.
%     It is obviously different from the original matrix, which is more flexible and not constrained to the physical connections of the human body.

% 	Fig.\ref{fig:layer357} shows an example of edges connected to $25_{th}$ joint in NTU-RGBD dataset learned by our model. 
%     Each circle represents one joint, whose size indicates the strength of the connection between current joint and the $25_{th}$ joint. 
%     The examples shown from left to right are the visualizations of the second subset of $3_{th}$, $5_{th}$ and $7_{th}$ layers in Fig.~\ref{fig:stream}, respectively. 
%     It shows that the connections and their strength is different for different layers. 
%     It verified our point that different layers contain different-level semantic information, which should own distinct graphs.
    
\begin{figure}[!htb]
	\begin{center}
	\includegraphics[width=0.7\linewidth]{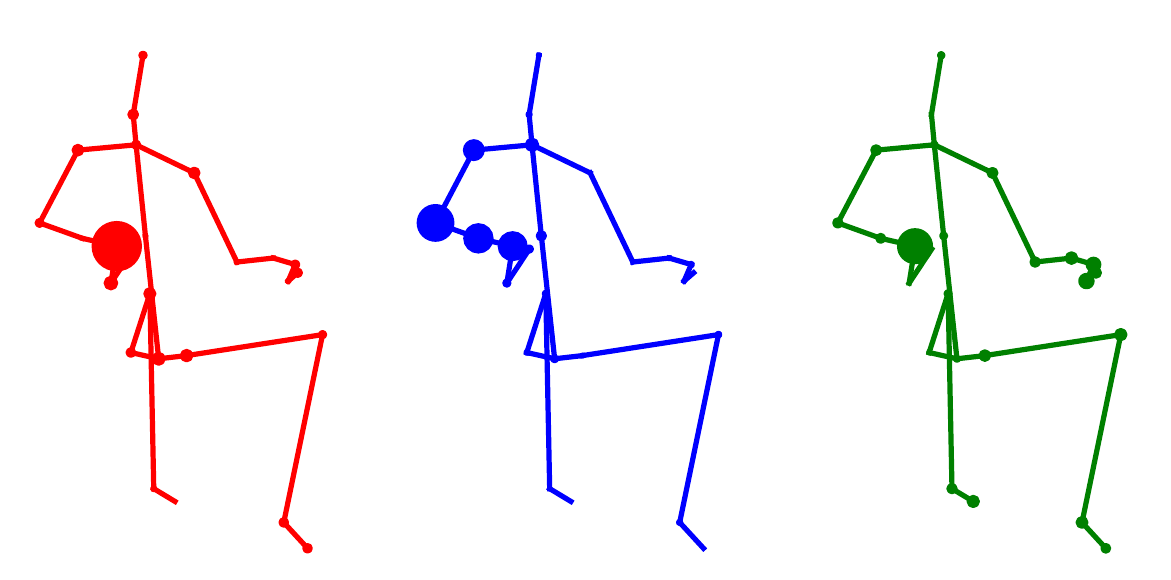}
	\caption{
    Visualization of the graphs for different layers.
%     Visualization of the edges for $25_{th}$ joint of right sketch in Fig.\ref{fig:dataset}. The size of the circle represents the tightness of the connection. From left to right is the visualization of different layers ($3_{th}$, $5_{th}$ and $7_{th}$ layer in Fig.~\ref{fig:stream}).
    }
	\label{fig:layer357}	
	\end{center}
	\end{figure}

%     Fig.~\ref{fig:layer357} is a visualization of the connections between other joints to the joint of right wrist ($25_{th}$ joint in NTU-RGBD dataset) in the learned adaptive graph of our model. 
    Fig.~\ref{fig:layer357} is a visualization of the skeleton graph for different layers of one sample (from left to right is the $3_{rd}$, $5_{th}$ and $7_{th}$ layers in Fig.~\ref{fig:stream}, respectively).
    The skeletons are plotted based on the physical connections of the human body.
    Each circle represents one joint, and its size represents the strength of the connection between the current  joint and the $25_{th}$ joint in the learned adaptive graph of our model. 
%     From left to right is the visualization of different layers (the $3_{th}$, $5_{th}$ and $7_{th}$ layer in Fig.~\ref{fig:stream}, respectively). 
    It shows that a traditional physical connection of the human body is not the best choice for the action recognition task, and different layers need graphs with different topology structures.
    The skeleton graph in the $3_{rd}$ layer pays more attention to the adjacent joints in the physical graph. This result is intuitive since the lower layer only contains the low-level feature, while the global information cannot be observed. 
    For the $5_{th}$ layer, more joints along the same arm are strongly connected. 
    For the $7_{th}$ layer, the left hand and the right hand show a stronger connection, although they are far away from each other in the physical structure of the human body. 
    We argue that a higher layer contains higher-level information. Hence, the graph is more relevant to the final classification task.
%     For example, the new learned graph of third skeleton pays more attention on the relation between left hand and right hand. 
%     In addition, different layer needs graphs with different topology structures since the information contained in each layer may alter.
    
\begin{figure}[!htb]
	\begin{center}
	\includegraphics[width=0.7\linewidth]{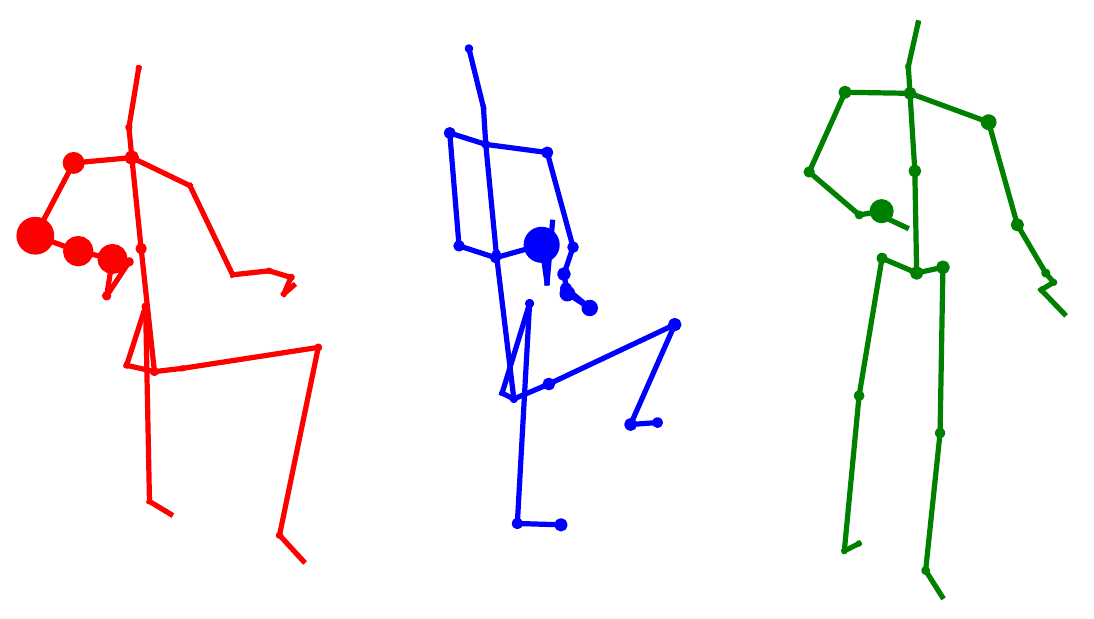}
	\caption{Visualization of the graphs for different samples.}
	\label{fig:sample012}
	\end{center}
	\end{figure}

    Fig.~\ref{fig:sample012} shows a similar visualization of Fig.~\ref{fig:layer357} but for different samples. 
    The learned adjacency matrix is extracted from the second subset of the $5_{th}$ layer in the model (Fig.~\ref{fig:stream}). 
    It shows that the graph structures learned by our model for different samples are also different, even for the same convolutional subset and the same layer. 
    It verified our point of view that different samples need different topologies of the graph, and a data-driven graph structure is better than a fixed one.
        
    \subsubsection{Two-stream framework}
    Another important improvement is the utilization of second-order information. 
    Here, we compare the performance of using each type of input data alone, shown as Js-AGCN and Bs-AGCN in Tab.~\ref{tab:twostream}, and the performance when combining them as described in Section~\ref{tsn}, shown as 2s-AGCN in Tab.~\ref{tab:twostream}.
    Clearly, the two-stream method outperforms the one-stream-based methods.

    \begin{table}[htb]
		\begin{center}
		\begin{tabular}{lc}
			\hline
			Methods     & Accuracy ($\%$)    \\
			\hline
			Js-AGCN & 93.7      \\
			Bs-AGCN & 93.2       \\
            \hline
            2s-AGCN & 95.1        \\
%             2s-NLGCN b & 94.0        \\
%             2s-NLGCN c & 95.1        \\
%             2s-NLGCN d & 95.0        \\
			\hline
		\end{tabular}
		\end{center}
        \caption{Comparisons of the validation accuracy with different input modalities.}
		\label{tab:twostream}
	\end{table}

    \subsection{Comparison with the state-of-the-art}
    We compare the final model with the state-of-the-art skeleton-based action recognition methods on both the NTU-RGBD dataset and Kinetics-Skeleton dataset. 
    The results of these two comparisons are shown in Tab~\ref{tab:ntu-rgbd} and Tab~\ref{tag:kinetics}, respectively. 
    The methods used for comparison include the handcraft-feature-based methods~\cite{vemulapalli_human_2014,fernando_modeling_2015}, RNN-based methods~\cite{du_hierarchical_2015,shahroudy_ntu_2016,liu_spatio-temporal_2016,song_end--end_2017,zhang_view_2017,li_skeleton-based_2018,li_independently_2018}, CNN-based methods~\cite{liu_two-stream_2017,kim_interpretable_2017,ke_new_2017,liu_enhanced_2017,li_skeleton-based_2017,li_skeleton_2017} and GCN-based methods~\cite{yan_spatial_2018,tang_deep_2018}.
    Our model achieves state-of-the-art performance with a large margin on both of the datasets, which verifies the superiority of our model.
%     \subsubsection{NTU-RGBD dataset}
%     In NTU-RGBD dataset, our model is compared with 
%     one hand-craft feature based methods, i.e. Lie Group \cite{vemulapalli_human_2014}, 
%     four LSTM based methods, i.e. HBRNN \cite{du_hierarchical_2015}, STA-LSTM \cite{song_end--end_2017}, VA-LSTM \cite{zhang_view_2017} and ARRN-LSTM~\cite{li_skeleton-based_2018}, 
%     four CNN based methods, i.e. TCN \cite{kim_interpretable_2017}, Synthesized CNN \cite{liu_enhanced_2017}, Motion+Trans+CNN \cite{li_skeleton-based_2017}, 3scale ResNet152 \cite{li_skeleton_2017},  
%     and one graph convolution based method, i.e. ST-GCN \cite{yan_spatial_2018}. 
%     These methods are briefly introduced in Section \ref{sec:relatedwork}. 
%     Tab.~\ref{ntu-rgbd} shows that the performance of deep learning based methods is generally better than hand-craft feature based methods, and CNN based methods are generally better than RNN based methods. 
%     Our model outperforms these methods with a large margin, which verifies the superiority of our model for skeleton-based action recognition.
    
      \begin{table}[htb]
      \begin{center}
		\begin{tabular}{lcc}
			\hline
			Methods     &X-Sub (\%)& X-View (\%)    \\
            \hline
            Lie Group \cite{vemulapalli_human_2014} & 50.1 & 82.8 \\
			\hline
            HBRNN~\cite{du_hierarchical_2015} & 59.1  &    64.0  \\
            Deep LSTM~\cite{shahroudy_ntu_2016}   & 60.7 &    67.3  \\
            ST-LSTM~\cite{liu_spatio-temporal_2016} & 69.2 & 77.7 \\
            STA-LSTM~\cite{song_end--end_2017}  & 73.4 &    81.2  \\
            VA-LSTM~\cite{zhang_view_2017}  & 79.2 &    87.7  \\
            ARRN-LSTM~\cite{li_skeleton-based_2018} & 80.7 & 88.8 \\
            Ind-RNN~\cite{li_independently_2018} & 81.8 & 88.0 \\
            \hline
            Two-Stream 3DCNN~\cite{liu_two-stream_2017} & 66.8 & 72.6 \\
            TCN~\cite{kim_interpretable_2017}   & 74.3&    83.1      \\
            Clips+CNN+MTLN~\cite{ke_new_2017} & 79.6 & 84.8 \\
            Synthesized CNN~\cite{liu_enhanced_2017}   & 80.0 &      87.2  \\
            CNN+Motion+Trans~\cite{li_skeleton-based_2017} & 83.2  &      89.3  \\
            3scale ResNet152~\cite{li_skeleton_2017}  & 85.0  &      92.3  \\
            \hline
            ST-GCN \cite{yan_spatial_2018} & 81.5  &      88.3  \\
            DPRL+GCNN~\cite{tang_deep_2018} & 83.5 & 89.8 \\
            \hline
            2s-AGCN (ours)&  88.5 & 95.1 \\
			\hline
		\end{tabular}
      \end{center}
      \caption{Comparisons of the validation accuracy with state-of-the-art methods on the NTU-RGBD dataset.}
      \label{tab:ntu-rgbd}
	\end{table}
    
%     \subsubsection{Kinetics-Skeleton dataset}
%     In Kinetics-Skeleton, our model is compared with 
%     one hand-crafted features based method, i.e. "Feature Encoding" \cite{fernando_modeling_2015}, 
%     one LSTM base methods, i.e. Deep LSTM \cite{shahroudy_ntu_2016}, 
%     one CNN based methods, i.e. TCN \cite{kim_interpretable_2017} and 
%     one graph convolution based method, i.e. ST-GCN \cite{yan_spatial_2018}. 
%     These methods are briefly introduced in Section \ref{sec:relatedwork}. The top-1 and top-5 recognition accuracies are reported in Tab.~\ref{kinetics}, where our model outperforms the other methods with a large margin.
    
    \begin{table}[htb]
    \begin{center}
		\begin{tabular}{lcc}
			\hline
			Methods     & Top-1 (\%)& Top-5 (\%)    \\
			\hline
			Feature Enc. \cite{fernando_modeling_2015} & 14.9  &    25.8  \\
			Deep LSTM  \cite{shahroudy_ntu_2016}   & 16.4 &    35.3  \\
			TCN  \cite{kim_interpretable_2017}   & 20.3&    40.0      \\
            ST-GCN  \cite{yan_spatial_2018}  & 30.7 &      52.8  \\
            \hline
            Js-AGCN (ours)& 35.1& 57.1\\
            Bs-AGCN (ours)& 33.3& 55.7\\
            2s-AGCN (ours)&  36.1& 58.7\\
			\hline
		\end{tabular}
    \end{center}
	\caption{Comparisons of the validation accuracy with state-of-the-art methods on the Kinetics-Skeleton dataset.}
    \label{tag:kinetics}
	\end{table}

	\subsection{Conclusion}
    In this work, we propose a novel adaptive graph convolutional neural network (2s-AGCN) for skeleton-based action recognition. 
    It parameterizes the graph structure of the skeleton data and embeds it into the network to be jointly learned and updated with the model. 
    This data-driven approach increases the flexibility of the graph convolutional network and is more suitable for the action recognition task. 
Furthermore, the traditional methods always ignore or underestimate the importance of second-order information of skeleton data, i.e., the bone information. 
    In this work, we propose a two-stream framework to explicitly employ this type of information, which further enhances the performance.
    The final model is evaluated on two large-scale action recognition datasets, NTU-RGBD and Kinetics, and it achieves the state-of-the-art performance on both of them.

\section*{Acknowledgement}
This work was supported in part by the National Natural Science Foundation of China under Grant 61572500, 61876182 and 61872364, and in part by the State Grid Corporation Science and Technology Project.
    
{\small
\bibliographystyle{ieee}
\bibliography{Zotero}
}

\end{document}